\documentclass[sigconf]{acmart}
\AtBeginDocument{%
  }

\copyrightyear{2025}
\acmYear{2025}
\setcopyright{acmlicensed}\acmConference[SIGIR '25]{Proceedings of the 48th International ACM SIGIR Conference on Research and Development in Information Retrieval}{July 13--18, 2025}{Padua, Italy}
\acmBooktitle{Proceedings of the 48th International ACM SIGIR Conference on Research and Development in Information Retrieval (SIGIR '25), July 13--18, 2025, Padua, Italy}
\acmDOI{10.1145/3726302.3730241}
\acmISBN{979-8-4007-1592-1/2025/07}

\begin{document}

\title{ReCDAP: Relation-Based Conditional Diffusion with Attention Pooling for Few-Shot Knowledge Graph Completion}

\author{Jeongho Kim}
\orcid{0009-0002-5409-7728}
\affiliation{
  \institution{Myongji University}
  \city{Yongin}
  \country{South Korea}}
\email{jeongho@mju.ac.kr}

\author{Chanyeong Heo}
\orcid{0009-0008-5884-4720}
\affiliation{
  \institution{Myongji University}
  \city{Yongin}
  \country{South Korea}}
\email{hcn98@mju.ac.kr}

\author{Jaehee Jung}
\orcid{0000-0002-0932-3039}
\authornote{Corresponding author.}
\affiliation{
  \institution{Myongji University}
  \city{Yongin}
  \country{South Korea}}
\email{jhjung@mju.ac.kr}

\renewcommand{\shortauthors}{Jeongho Kim, Chanyeong Heo, \& Jaehee Jung}

\begin{abstract}
Knowledge Graphs (KGs), composed of triples in the form of (head, relation, tail) and consisting of entities and relations, play a key role in information retrieval systems such as question answering, entity search, and recommendation. In real-world KGs, although many entities exist, the relations exhibit a long-tail distribution, which can hinder information retrieval performance. 
Previous few-shot knowledge graph completion studies focused exclusively on the positive triple information that exists in the graph or, when negative triples were incorporated, used them merely as a signal to indicate incorrect triples. To overcome this limitation, we propose Relation-Based Conditional Diffusion with Attention Pooling (ReCDAP). First, negative triples are generated by randomly replacing the tail entity in the support set. By conditionally incorporating positive information in the KG and non-existent negative information into the diffusion process, the model separately estimates the latent distributions for positive and negative relations. Moreover, including an attention pooler enables the model to leverage the differences between positive and negative cases explicitly. Experiments on two widely used datasets demonstrate that our method outperforms existing approaches, achieving state-of-the-art performance. The code is available at https://github.com/hou27/ReCDAP-FKGC.
\end{abstract}




\maketitle

\section{Introduction}
A knowledge graph (KG) structures facts contained in a knowledge base in the form of (head, relation, tail) \cite{yago, nell, wiki}. It is employed in various information retrieval systems such as recommendation \cite{graph-based_recommender_systems}, search engines \cite{search_engine_assistant}, and question answering \cite{knowledge_graph_question_answering}. In the real world, knowledge graphs are incomplete, and relation data exhibit a long-tail distribution. So, the few-shot knowledge graph completion (FKGC) task, which predicts a missing tail entity based on limited information, is an important research area.

Few-shot knowledge graph completion approaches can be broadly divided into metric learning-based and meta-learning-based models. Metric learning-based models capture the similarity between query and support triples in the embedding space. For example, GMatching \cite{xiong-etal-2018-one} generates entity embeddings using local graph information and an LSTM-based ranking module. FSRL \cite{zhang2020few} employs an attention-based neighbor encoder with fixed weights, while FAAN \cite{sheng-etal-2020-adaptive} dynamically weights neighbors based on the relation. CIAN \cite{li-etal-2022-learning-inter} aggregates neighbor information in two stages. In contrast, meta-learning-based models update parameters via the support set to predict the tail entity. For instance, MetaR \cite{chen-etal-2019-meta} adjusts relation meta parameters using loss gradients, and GANA \cite{gana2021} mitigates neighbor noise with gating and attention while modeling relations using a TransH-based approach. NP-FKGC \cite{npfkgc2023} combines a normalizing flow with a neural process to model distributions that reconcile differences between the support and query sets, computing triple scores via a probabilistic decoder. Some studies \cite{chen-etal-2019-meta, gana2021, npfkgc2023} additionally constructed a support negative set by replacing the correct tail entity in the support set with an incorrect entity, thereby utilizing positive and negative information. However, since it was used solely as a signal indicating an incorrect triple, it was not possible to accurately model the distributional properties or relation-specific characteristics inherent in the negative data.

Diffusion, a probabilistic generative model, has recently been applied to knowledge graph tasks \cite{jiang2024diffkg, dkgc2024, long2024kgdm, niu2025diffusion}. In particular, the Fact Embedding through Diffusion Model (FDM) \cite{fdm2024www} learns the distribution of knowledge graph facts by reframing entity prediction as conditional fact generation. Leveraging the Denoising Diffusion Probabilistic Model (DDPM) \cite{ho2020denoising}, it captures complex relational patterns and improves knowledge graph completion performance.

This study proposes a novel approach that explicitly models positive and negative information through a diffusion model. Instead of using support negative triples merely as a contrastive signal, our model leverages relation, positive/negative representations, and label information to accurately estimate separate distributions for positive and negative cases. The attention pooler extracts key features from each distribution to ensure effective utilization of negative information in tail entity prediction. Experiments show that our method outperforms state-of-the-art approaches on NELL and FB15K.
The main contributions of this paper are summarized as follows:
\begin{itemize} 
\item A novel diffusion-based modeling approach is proposed in which the diffusion process is conditioned on relation-based information, thereby explicitly separating and accurately estimating the positive and negative embedding distributions.
\item An attention pooler is employed to independently extract key features from both distributions, ensuring that negative information is effectively utilized in tail entity prediction.
\item Experiments confirm that our method outperforms existing approaches on benchmark datasets. 
\end{itemize}

\begin{figure*}[ht]
  \centering
  \includegraphics[width=\linewidth]{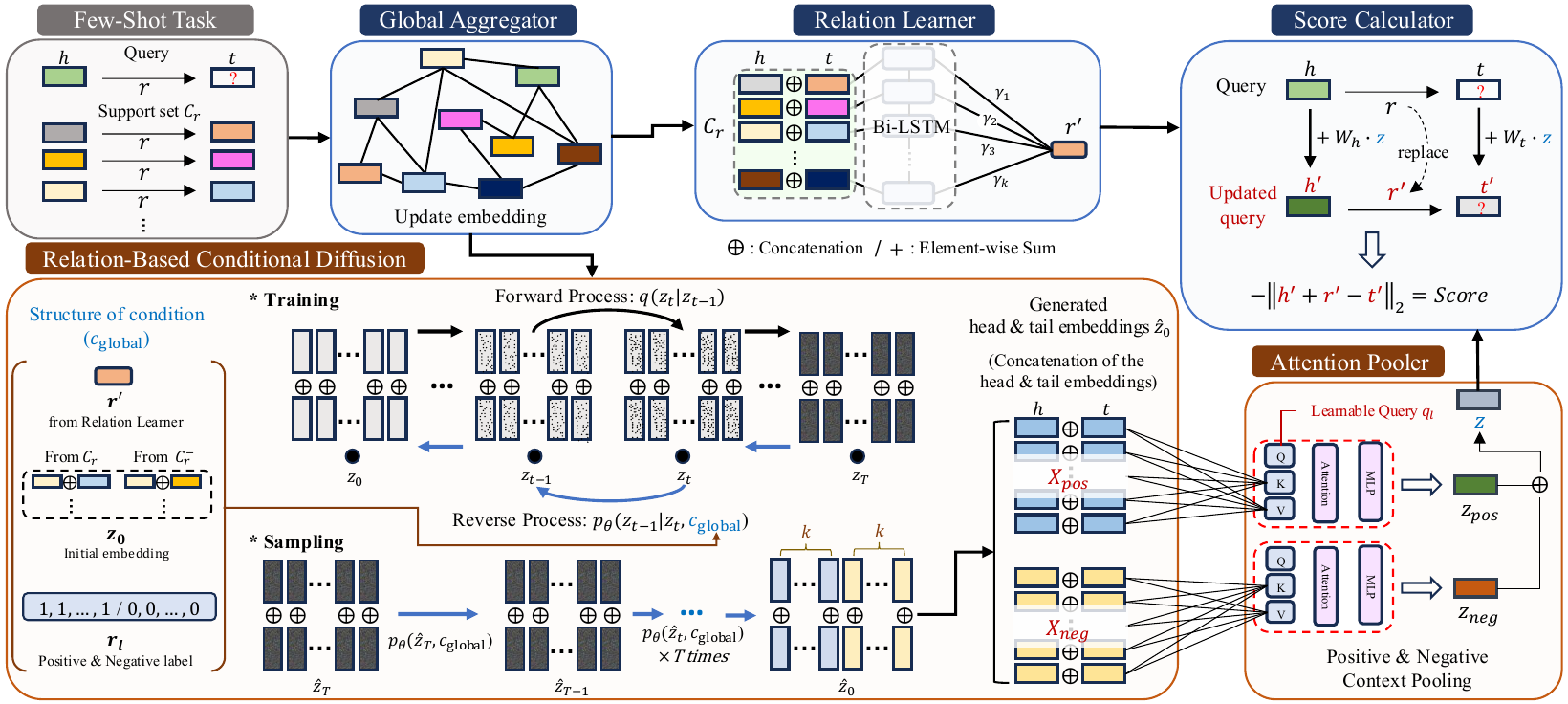}
  \caption{Model overview}
  \label{fig:overview}
  \Description{model process}
\end{figure*}

\section{Methodology}
Figure \ref{fig:overview} illustrates our model's five components: a global aggregator, relation learner, relation-based conditional diffusion (ReCD), attention pooler, and score calculator. The global aggregator updates embeddings and the relation learner computes the relation $r'$ from the support set. The ReCD module then estimates positive and negative distributions from the extended support set $\tilde{C}_r$ (formed by appending a support negative set $C^-_r$ generated by replacing the tail entity with a random negative entity to the original support set $C_r$). Subsequently, the attention pooler extracts the latent representation $z$. Finally, the score calculator computes triple scores.

\subsection{Global Aggregator}
In this work, the GNN module proposed in NP-FKGC \cite{npfkgc2023} is employed. 
The embeddings of the head and tail entities for all triples are computed as the weighted sum of neighbors up to 2-hop based on the relation. Each edge $(u,v)$ produces a message via $m_{uv} = W_r\, [h_u; \, e_{uv}; \, h_v]$, where $[h_u; \, e_{uv}; \, h_v]$ denotes the concatenation of the source node feature $h_u$, the edge feature $e_{uv}$, and the target node feature $h_v$. The attention coefficient is computed as
\begin{equation}
\alpha_{uv} = \frac{\exp\Bigl(\mathrm{LeakyReLU}\bigl(w^\top [h_v; \, m_{uv}]\bigr)\Bigr)}{\sum_{u'\in\mathcal{N}(v)} \exp\Bigl(\mathrm{LeakyReLU}\bigl(w^\top [h_v; \, m_{u'v}]\bigr)\Bigr)}.
\end{equation}
Here, $\mathcal{N}(v)$ denotes the set of neighbors of node $v$. Then the node feature is updated by
\begin{equation}
h'_v = \sum_{u\in\mathcal{N}(v)} \alpha_{uv}\, m_{uv} + W_{\text{loop}}\, h_v.
\end{equation}
Updated node embeddings $h'_v$ are directly used in subsequent modules. For each triple in the extended support set $\tilde{C}_r$, the head and tail embeddings are concatenated to form the triple's representation $z_0$.

\subsection{Relation Learner}
The module for computing the relation representation adopts the same structure as the Bi-LSTM relation learner used in previous studies \cite{gana2021, npfkgc2023}. The support set $C_r$ is treated as an input sequence and processed by a Bi-LSTM to obtain $\{h_t\}_{t=1}^{k}$, where $k$ denotes the number of support triples. Then, the relation representation $r'$ is computed via an attention mechanism as
\begin{equation}\label{eq:relation_learner}
\begin{aligned}
r' = W_{\mathrm{out}}\left(\sum_{t=1}^{k} \gamma_t\, h_t\right),\quad
\gamma_t = \frac{\exp\bigl(w^\top h_t\bigr)}{\sum_{t=1}^{k} \exp\bigl(w^\top h_t\bigr)}.
\end{aligned}
\end{equation}

\subsection{Relation-Based Conditional Diffusion with Attention Pooling}
In this section, the following two components are introduced to jointly model and disentangle positive and negative distributions.
\subsubsection{Relation-Based Conditional Diffusion}
In this study, the diffusion model progressively adds noise to the original embedding $z_0$ following a Markov chain, as described in Eq.~\ref{eq:diffusion1}. Next, as shown in Eq.~\ref{eq:diffusion2}, the model approximates the original data distribution $q(z_0)$ by predicting the added noise conditioned on relation-based information $c_{\text{global}}$. Finally, as outlined in Eq.~\ref{eq:diffusion3}, new data is generated by sampling from the learned distribution $p_\theta(z_0)$ \cite{sohl2015deep, ho2020denoising}. Note that $\beta_t$ denotes the variance, $\alpha_t = 1 - \beta_t$ and $\bar{\alpha}_t = \prod_{s=1}^{t} \alpha_s$. 

\begin{equation}\label{eq:diffusion1}
q(z_t \mid z_{t-1}) = \mathcal{N}\Bigl( z_t; \sqrt{1-\beta_t}\,z_{t-1},\,\beta_t\,\mathbf{I} \Bigr)
\end{equation}

\begin{equation}\label{eq:diffusion2}
\begin{aligned}
p_{\theta}(z_{t-1} \mid z_t, c_{\text{global}}) := \mathcal{N}\left( z_{t-1}; \mu_{\theta}(z_t, t, c_{\text{global}}), \Sigma_{\theta}(z_t, t, c_{\text{global}}) \right),\\
\text{where}\quad \mu_{\theta}(z_t, t, c_{\text{global}}) = \frac{1}{\sqrt{\alpha_t}} \Bigl( z_t - \frac{\beta_t}{\sqrt{1-\bar{\alpha}_t}}\,\epsilon_{\theta}(z_t, t, c_{\text{global}}) \Bigr).
\end{aligned}
\end{equation}

\begin{equation}\label{eq:diffusion3}
p_\theta(z_{0:T}) = p(z_T) \prod_{t=1}^{T} p_\theta(z_{t-1}\mid z_t, c_{\text{global}})
\end{equation}

The diffusion model operates as follows. In the forward process, noise is progressively added to the embedding $z_0$ of the head and tail entities in the original triple based on the forward process variances $\beta_t$, transforming it into $z_1,\, z_2,\, \dots,\, z_T$, whereby $q(z_T \mid z_0)$ converges to a complete Gaussian distribution. Subsequently, in the reverse process, starting from the $z_T$, the model denoises to recover the original embedding $z_0$. Importantly, rather than merely denoising, the model is configured to accurately distinguish the distributional differences between positive and negative cases by conditioning on additional information. The condition is formed by the concatenation of three pieces of information as $c_{\text{global}} = \mathrm{Concat}\left(r',\, z_0,\, r_l\right)$. Through the relation information $r'$ and the triple embedding $z_0$, which is constructed by concatenating the updated head and tail embeddings from the extended support set, the model captures the positive and negative distributions of the relation. Moreover, the label $r_l$, which explicitly indicates whether each triple represented by $z_0$ corresponds to a positive or negative triple, provides clear guidance for accurately distinguishing between these distributions during noise removal. In the sampling process, starting from the initial noisy representation $\hat{z}_T$, the learned model denoises to generate the denoised embedding $\hat{z}_0$.
A U-Net structured model is employed to predict the added noise $\epsilon_{\theta}(z_t, t, c_{\text{global}})$ defined in Eq.~\ref{eq:diffusion2}, with the FiLM \cite{perez2018film} technique applied to incorporate $c_{\text{global}}$ as a conditioning factor.

\subsubsection{Attention Pooler}
The attention pooler is based on the transformer \cite{NIPS2017_3f5ee243} architecture that separates positive and negative information and performs pooling using a single query for each, thereby capturing the important features in the data more effectively than mean or max pooling. The attention pooler extracts the embedding of the context by performing an attention operation with a learnable query $q_l$ on the input $X$ and then outputs the embedding through an MLP. 
In a $k$-shot setting, $\hat{z}_0$ is split into $X_{pos}$ (first $k$ elements) and $X_{neg}$ (remaining). As shown in Eq.~\ref{eq:ap}, the attention pooler produces $z_{\text{pos}}$ and $z_{\text{neg}}$, respectively.

\begin{equation}\label{eq:ap}
\begin{aligned}
\mathrm{AttnPool}(X) = \mathrm{MLP}\bigl(&\mathrm{MultiHeadAttention}(q_l, X, X)\bigr) \\
z_{\text{pos}} = \mathrm{AttnPool}&(X_{\text{pos}}), \quad
z_{\text{neg}} = \mathrm{AttnPool}(X_{\text{neg}})
\end{aligned}
\end{equation}
and the final latent representation $z$ is obtained by concatenation:
\begin{equation}
z = [z_{\text{pos}}; \, z_{\text{neg}}].
\end{equation}

\subsection{Score Calculator}
The computation of the tail entity score for a given query is based on the TransE \cite{NIPS2013_1cecc7a7} formulation. As shown in Eq. \ref{eq:project_entities_and_scoring}, the embeddings of the head and tail entities are updated using the $z$ from the previous stage. Then, the final score is computed. By leveraging $z$, the model incorporates relation-conditioned positive and negative information, resulting in a more refined score estimation.

\begin{equation}\label{eq:project_entities_and_scoring}
    h' = h + W_h\, z, \quad t' = t + W_t\, z, \quad \text{Score} = -\left\| h' + r' - t' \right\|_2. 
\end{equation}

\section{Optimization}
We minimize a composite loss consisting of two terms. The first is a margin ranking loss that ensures a positive triple's score exceeds a negative triple's score by at least a margin $\delta$. Let $\{p_i\}_{i=1}^N$ and $\{n_i\}_{i=1}^N$ denote the positive and negative scores, respectively, where $N$ is the number of queries. Formally,
\begin{equation}
\mathcal{L}_{\text{margin}} = \frac{1}{N}\sum_{i=1}^{N} \max\{0,\, \delta - (p_i - n_i)\}.
\end{equation}

The second term is the MSE loss \cite{ho2020denoising} used to train the diffusion module. A noisy sample $z_{\text{noisy}}$ is generated by adding noise to the original embedding $z_0$ over $t$ steps. The loss is then computed as the mean squared error between the model-predicted noise under the condition $c_{\text{global}}$, denoted as $\epsilon_\theta(z_{\text{noisy}}, t, c_{\text{global}})$, and the noise $\epsilon$ sampled from a Gaussian distribution:
\begin{equation}
\mathcal{L}_{\text{MSE}} = \frac{1}{N}\sum_{i=1}^{N} \left\|\epsilon_\theta(z_{\text{noisy}}, t, c_{\text{global}}) - \epsilon\right\|_2^2.
\end{equation}

Thus, the total loss is given by:
\begin{equation}
\mathcal{L}_{\text{total}} = \mathcal{L}_{\text{margin}} + \mathcal{L}_{\text{MSE}}.
\end{equation}

\section{Experiment}

\begin{table}[h]
  \caption{Statistics of the datasets}
  \label{tab:dataset_statistics}
  \resizebox{\columnwidth}{!}{
  \begin{tabular}{ccccccc}
    \toprule
    \textbf{Dataset} & \textbf{\#Entity} & \textbf{\#Triples} & \textbf{\#Relation} & \textbf{\#Train} & \textbf{\#Valid} & \textbf{\#Test} \\
    \midrule
    NELL      & 68,545  & 181,109  & 358  & 51  & 5   & 11  \\
    FB15K-237 & 14,541  & 281,624  & 231  & 75  & 11  & 33  \\
    \bottomrule
  \end{tabular}}
\end{table}

\begin{table*}[ht]
  \caption{Performance Comparison on NELL and FB15K-237}
  \label{tab:results}
  \begin{tabular}{c|cccc|cccc}
    \toprule
    \textbf{Methods} & \multicolumn{4}{c|}{\textbf{NELL}} & \multicolumn{4}{c}{\textbf{FB15K-237}} \\
    \cmidrule(lr){2-5} \cmidrule(lr){6-9}
                     & \textbf{MRR} & \textbf{Hits@10} & \textbf{Hits@5} & \textbf{Hits@1} 
                     & \textbf{MRR} & \textbf{Hits@10} & \textbf{Hits@5} & \textbf{Hits@1} \\
    \midrule
    GMatching        & 0.176 & 0.294 & 0.233 & 0.113  & 0.325* & 0.509* & 0.426* & 0.233* \\
    MetaR            & 0.261 & 0.437 & 0.350 & 0.168  & 0.429* & 0.595* & 0.524* & 0.342* \\
    FSRL             & 0.154* & 0.210* & 0.159* & 0.128*  & 0.331* & 0.436* & 0.380* & 0.272* \\
    FAAN             & 0.279 & 0.428 & 0.364 & 0.200  & 0.416* & 0.592* & 0.508* & 0.328* \\
    GANA             & 0.344 & 0.517 & 0.437 & 0.246  & 0.447* & 0.654* & 0.574* & 0.338* \\
    CIAN             & 0.376 & 0.527 & 0.453 & 0.298  & 0.482* & 0.651* & 0.577* & 0.393* \\
    NP-FKGC          & 0.460 & 0.494 & 0.471 & 0.437  & 0.538 & 0.671 & 0.593 & 0.476 \\
    \textbf{ReCDAP} & \textbf{0.505} & \textbf{0.528} & \textbf{0.506} & \textbf{0.493} 
                     & \textbf{0.579} & \textbf{0.745} & \textbf{0.695} & \textbf{0.491} \\
    \bottomrule
  \end{tabular}
  \parbox{0.7\linewidth}{\footnotesize{* denotes results obtained using the authors' official implementation. MRR denotes Mean Reciprocal Rank. Hits@N denotes the proportion of cases in which the correct tail entity is ranked within the top $N$.}}
\end{table*}

\subsection{Datasets}
The performance of the proposed method was compared against other baselines on two public benchmark datasets (NELL, FB15K-237) \cite{xiong-etal-2018-one, npfkgc2023, wang2021reform}. The NELL dataset was split into train/validation/test sets in the ratio of 51/5/11, and FB15K-237 was split in the ratio of 75/11/13. The statistics for the two datasets are provided in Table \ref{tab:dataset_statistics}.

\subsection{Results}
Few-shot knowledge graph completion (FKGC) aims to identify the correct tail entity for a given $\text{query} = (h, r, ?)$ when provided with a specific relation $r$ and $n$ corresponding positive triples. To evaluate the effectiveness of our method, we compared performance with seven previous FKGC methods: GMatching \cite{xiong-etal-2018-one}, MetaR \cite{chen-etal-2019-meta}, FSRL \cite{zhang2020few}, FAAN \cite{sheng-etal-2020-adaptive}, GANA \cite{gana2021}, CIAN \cite{li-etal-2022-learning-inter}, and NP-FKGC \cite{npfkgc2023}. To ensure fairness, we used the results reported in those studies without re-running. Entity and relation embeddings were initialized with pre-trained TransE embeddings from GMatching \cite{xiong-etal-2018-one} and REFORM \cite{wang2021reform}. For the proposed ReCDAP, the diffusion iteration step was set to 100. To ensure valid shape transitions during upsampling and downsampling within the diffusion model, zero-padding was applied to extend the sequence length beyond a minimum threshold. The required padding length depends on the number of support triples. The attention pooler was configured with a single head. The attention pooler was configured with a single head. For both NELL and FB15K-237, the embedding dimension was set to 100. The learning rate was set to $1 \times 10^{-3}$ for NELL and $1 \times 10^{-4}$ for FB15K-237, and the margin value was set to 1 in all cases. The batch size was 64 for NELL and 32 for FB15K-237. All experiments were conducted on a single NVIDIA RTX 3090 GPU (24GB RAM). The experimental results under the 5-shot setting, as shown in Table \ref{tab:results}, demonstrate that state-of-the-art performance was achieved on both datasets compared to existing models. 
While NP-FKGC focuses on modeling differences between the support and query sets, our approach recognizes that support triples sharing the same relation can exhibit a multi-modal distribution. Thus, we directly estimate its distribution via conditional diffusion. Furthermore, rather than merely employing a support negative set, our method uses relation-based conditional diffusion with attention pooling to explicitly model both positive and negative distributions. This leads to more effective tail prediction and improved overall performance.

\subsection{Ablation Study}

\subsubsection{Relation-Based Conditional Diffusion and Attention Pooler}
Experiments were conducted to validate the roles of the relation-based conditional diffusion and attention pooler. First, w/o ReCD refers to replacing the diffusion module with the MLP used in NP-FKGC, and w/o AttnPool replaces the attention pooler with mean pooling. Lastly, w/o ReCD \& AttnPool indicates that both components are removed. As shown in Table \ref{tab:ablation_study_recd_attnpool}, the best performance was achieved when both were used. Moreover, performance decreases in the order of using only relation-based conditional diffusion, then only the attention pooler, and finally removing both, thereby demonstrating that each component plays a critical role.

\begin{table}[h]
  \caption{Validation of ReCD and Attention Pooler on NELL}
  \label{tab:ablation_study_recd_attnpool}
  \resizebox{0.9\columnwidth}{!}{
  \begin{tabular}{c|cccc}
    \toprule
    \textbf{Variants} & \textbf{MRR} & \textbf{Hits@10} & \textbf{Hits@5} & \textbf{Hits@1} \\
    \midrule
    ReCDAP & \textbf{0.505} & \textbf{0.528} & \textbf{0.506} & \textbf{0.493} \\
    w/o ReCD & 0.440 & 0.456 & 0.440 & 0.427 \\
    w/o AttnPool & 0.449 & 0.452 & 0.452 & 0.441 \\
    w/o ReCD \& AttnPool & 0.281 & 0.303 & 0.285 & 0.265 \\
    \bottomrule
  \end{tabular}}
\end{table}

\subsubsection{Effect of Support Negative Information}
Two experimental settings were examined to evaluate the effect of incorporating negative information during triple score computation. First, w/o Support Negative Set indicates that the support negative set is entirely removed from the extended support set. Next, w/o Positive/Negative Separation refers to the scenario in which label information that explicitly distinguishes between positive and negative in the diffusion condition is not provided, and the attention pooler processes positive and negative information in an integrated manner. As shown in Table \ref{tab:ablation_study_posneg}, removing the support negative set results in a drop in performance, and notably, the configuration without explicit positive/negative separation exhibits an even more significant performance decline than the removal of the support negative set.

\begin{table}[h]
  \caption{Effect of Support Negative Information on NELL}
  \label{tab:ablation_study_posneg}
  \resizebox{\columnwidth}{!}{
  \begin{tabular}{c|cccc}
    \toprule
    \textbf{Variants} & \textbf{MRR} & \textbf{Hits@10} & \textbf{Hits@5} & \textbf{Hits@1} \\
    \midrule
    ReCDAP & \textbf{0.505} & \textbf{0.528} & \textbf{0.506} & \textbf{0.493} \\
    w/o Support Negative Set & 0.471 & 0.507 & 0.476 & 0.453 \\
    w/o Positive/Negative Separation & 0.454 & 0.455 & 0.448 & 0.444 \\
    \bottomrule
  \end{tabular}}
\end{table}

\section{Conclusion}
In this study, we proposed a model that effectively leverages negative information for the FKGC task by employing a diffusion module with relation-based conditioning and an attention pooler. The ablation study confirmed that the performance gains are due to two primary factors: the diffusion module's ability to capture distinct positive and negative distributions through relation-based information and the attention pooler's effective extraction of their key features. Our results highlight the critical role of explicitly integrating negative information in enhancing model performance.
Despite the diffusion module's strength in capturing complex distributions, its iterative nature introduces substantial computational overhead and slower inference speeds. Furthermore, aggregating multi-hop neighbors for high-degree datasets like Wiki \cite{wiki} remains memory-inefficient. Therefore, future work will focus on developing memory-efficient multi-hop aggregation techniques and designing lightweight models, which will be crucial for sustaining performance and scalability in large-scale deployments.

\begin{acks}
This work was supported by the National Research Foundation of Korea (NRF) grant funded by the Korean government (MSIT) (NRF-2022R1F1A1061476).
\end{acks}

\bibliographystyle{ACM-Reference-Format}
\balance
\bibliography{reference}

\end{document}